\documentclass[journal]{IEEEtran}
\usepackage{hyperref}
\usepackage{algorithm}
\usepackage{algorithmic}
\usepackage{graphicx}

\usepackage{algorithmic}

\usepackage{caption}

\usepackage[caption=false,font=footnotesize]{subfig}
\usepackage{array}

\usepackage{amsfonts}
\usepackage{mathtools}
\usepackage{amsmath}
\usepackage{comment}
\usepackage{verbatim}
\usepackage{textcomp}
\usepackage{multirow}
\usepackage[table,xcdraw]{xcolor}
\usepackage{wasysym} 

\usepackage{graphicx,dblfloatfix}

\usepackage{balance}

\usepackage{cite}
\usepackage{xcolor}
\usepackage[normalem]{ulem}

\usepackage{textcomp}
\usepackage{url}

\newcommand{\sectionref}[1]{Section\,\ref{#1}}
\newcommand{\figureref}[1]{Fig. \,\ref{#1}}
\newcommand{\tableref}[1]{Table\,\ref{#1}}

\title{CPG-Based Manipulation with Multi-Module Origami Robot Surface
}

\author{Yuhao Jiang$^{1}$, Serge El Asmar$^{1}$, Ziqiao Wang$^{1}$, Serhat Demirtas$^{1}$, and Jamie Paik$^{1}$
\thanks{This work is supported by the European Union's Horizon Europe Research and Innovation Programme under grant No.\,101069536. \textit{(Corresponding author: Jamie Paik)}}
\thanks{$^{1}$Reconfigurable Robotics Laboratory, School of Engineering, EPFL, Lausanne, 1005, Switzerland.  \{yuhao.jiang, serge.elasmar, ziqiao.wang, \\ serhat.demirtas, jamie.paik\}@epfl.ch}   
\thanks{Code available at: \url{https://github.com/DuxtX/Simulation_CPG_Manipulation_Oripixel.git} under Apache 2.0 License.}
}

\begin{document}

\maketitle

\begin{abstract}

Robotic manipulators often face challenges in handling objects of different sizes and materials, limiting their effectiveness in practical applications. This issue is particularly pronounced when manipulating meter-scale objects or those with varying stiffness, as traditional gripping techniques and strategies frequently prove inadequate. In this letter, we introduce a novel surface-based multi-module robotic manipulation framework that utilizes a Central Pattern Generator (CPG)-based motion generator, combined with a simulation-based optimization method to determine the optimal manipulation parameters for a multi-module origami robotic surface (Ori-Pixel). This approach allows for the manipulation of objects ranging from centimeters to meters in size, with varying stiffness and shape. The optimized CPG parameters are tested through both dynamic simulations and a series of prototype experiments involving a wide range of objects differing in size, weight, shape, and material, demonstrating robust manipulation capabilities.
\end{abstract}


\section{Introduction}
\label{sec:intro}

\begin{figure}[t]
    \centering
    \includegraphics[width = \columnwidth]{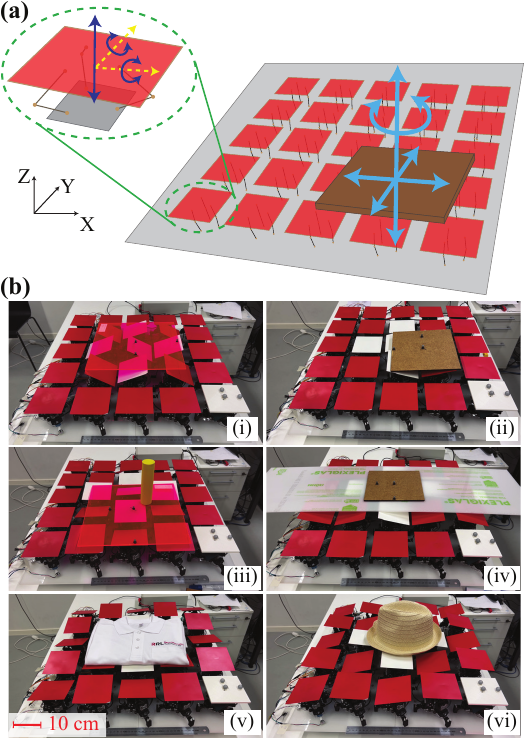}
    \caption{\textbf{Conceptual overview.} 
    (a) Conceptual illustration of the proposed 3-DoF CPG-based manipulation framework using a multi-module origami robot surface;
    (b) experiments demonstrating the system's versatility for manipulating various objects: 
    (i) 300$\times$300\,mm acrylic plate, 
    (ii) 200$\times$200\,mm wood plate, 
    (iii) 300$\times$300\,mm acrylic plate with a slender foam cylinder loosely positioned on top,
    (iv) 1000$\times$300\,mm acrylic plate weighing 1\,kg, 
    (v) 400$\times$400\,mm Polo shirt weighing 280\,g, and
    (vi) 250$\times$270\,mm Trilby hat weighing 55\,g.
    }
    \label{fig:concept}
\end{figure}

Robotic manipulation has made significant strides in recent years, leveraging advanced control and planning algorithms to demonstrate a variety of automated and precise tasks. Traditional robotic manipulators, typically employing robotic arms and grippers, have shown remarkable versatility in handling objects of different materials and shapes~\cite{shintake_versatile_2016, shintake_soft_2018, khurana_motion_2024}. When combined with advanced learning-based control strategies~\cite{cui_toward_2021}, these systems can perform intricate tasks such as in-hand manipulations~\cite{abondance_dexterous_2020, andrychowicz_learning_2020, liu_modeling_2020}, teleoperation~\cite{aldaco_aloha_nodate, 10035484}, dynamic stabilization~\cite{9811752}, dynamic throwing~\cite{liu_tube_2024}, and dressing~\cite{zhang_learning_2022}.

Traditional robotic grippers excel in their designed applications but often face scalability challenges when handling objects of varying types and sizes \cite{shintake2018soft}. For instance, ``Gripping by Actuation'' approaches effectively handle convex objects but show limitations with deformable materials \cite{crooks2016fin, crooks2017passive}. While controlled-stiffness grippers \cite{6225373,7086317} and grippers with integrated adhesion \cite{shintake2016,hawkes2015grasping} offer unique advantages for specific object types, achieving versatile manipulation remains an open challenge, particularly for objects at meter-scale and with diverse material properties.

To address these challenges, researchers have explored alternative approaches such as dynamic planar robotic surfaces. These surfaces, often using arrays of 1-DoF pins or more complex mechanisms like delta robots, have shown promise in manipulating various objects~\cite{leithinger_shape_2015, barr_smart_2013, follmer_inform_2013, xue_arraybot_2023, thompson_towards_2021, patil_linear_2023}. However, such systems usually require a significant number of actuators and sophisticated control methods, which limit their applications. Other novel actuators, including soft pneumatic actuators~\cite{deng_novel_2016, robertson_compact_2019}, ciliary actuators~\cite{ataka_design_2009}, and liquid crystal elastomers~\cite{liu_robotic_2021}, have also been investigated to address these limitations. Nevertheless, these approaches have yet to fully overcome the challenges posed by larger objects or flexible materials.

Central Pattern Generators (CPGs) have been extensively studied for generating locomotion in robots~\cite{ijspeert_central_2008}. By producing rhythmic signals, CPG-based controllers have proven effective in simplifying control requirements, thereby reducing actuation complexity in various multi-legged robotic systems, including bipedal robots~\cite{badri-sprowitz_birdbot_2022, 10499824}, quadrupedal robots~\cite{Cohen2003,cheetah2013,10175020}, and swimming robots~\cite{Porez2014,4459741}. Despite their prevalence in robotic locomotion, CPGs have seen limited application in robotic manipulation.

In this letter, we introduce a novel framework for manipulating objects of diverse sizes and stiffness, ranging from centimeters to meters, using the previously developed multi-module origami robotic surface - Ori-Pixel~\cite{Oripixel}. This approach combines a collective CPG-based manipulation motion generator with simulation-based optimizations. As shown in~\figureref{fig:concept}(a), our method utilizes the Canfield parallel origami robot, which offers three degrees of freedom: Z-axis translation and rotation around the X and Y axes. By arranging these robots in a 5$\times$5 multi-module array, we enable versatile manipulations including fast and smooth translations and rotations for objects of varying scales and stiffness.

The key challenge in controlling this platform lies in its high dimensionality, with 75 degrees of freedom (DoF) across the array. While this high-DoF configuration provides exceptional flexibility and precision for complex, localized manipulation tasks, it also presents significant challenges for control synthesis. Traditional control methods struggle with the complex kinematics and actuation coordination, while learning-based approaches face difficulties due to the vast search space, challenges in collecting comprehensive training data, and limited adaptability to hardware modifications. For instance, adding or removing a row of modules would typically necessitate complete model retraining in learning-based methods. To address these challenges, we introduce a CPG-based method that strategically groups the modules and represents end-effector motions using synchronized sinusoidal functions, effectively reducing the control optimization targets from 75 individual actuator positions to only 8 parameters. This reduction dramatically simplifies the optimization space, improving both the search efficiency for optimal control parameters and the system's real-world applicability. Moreover, our CPG-based approach offers inherent flexibility to platform modifications, as the underlying control principles remain valid regardless of the specific module configuration.

The proposed framework employs simulation-based optimization of the CPG parameters to generate effective motion patterns across the robotic surface. Through dynamic simulations and prototype experiments, we demonstrate the framework's capability to translate and rotate objects of varying sizes and materials, from rigid wood and acrylic to flexible fabrics. Fine-tuning objective functions allows our CPG-based controller to operate in two distinct modes - fast manipulation and smooth, stable manipulation - with the latter being particularly beneficial for delicate items sensitive to sudden positional and orientational changes, making it an adaptable solution for a broad spectrum of manipulation tasks.

The contributions of this letter are summarized as follows:
\begin{enumerate}
    \item  A novel CPG-based motion generator is developed for manipulations using multi-module robot surface, enabling various collective manipulation modes for objects of diverse sizes and stiffness.
    \item  A simulation-based optimization framework is then proposed to guide selecting optimal CPG parameters across a range of object settings and manipulation modes.
    \item  Dynamic simulations and prototype experiments are conducted to validate the proposed motion design and optimization framework, demonstrating effective manipulations of objects with varying sizes, shapes, and stiffness.
\end{enumerate}

\section{Kinematic Modeling and Dynamic Simulation}
\label{sec:modeling}
\begin{figure}[t]
    \centering
    \includegraphics[width = \columnwidth]{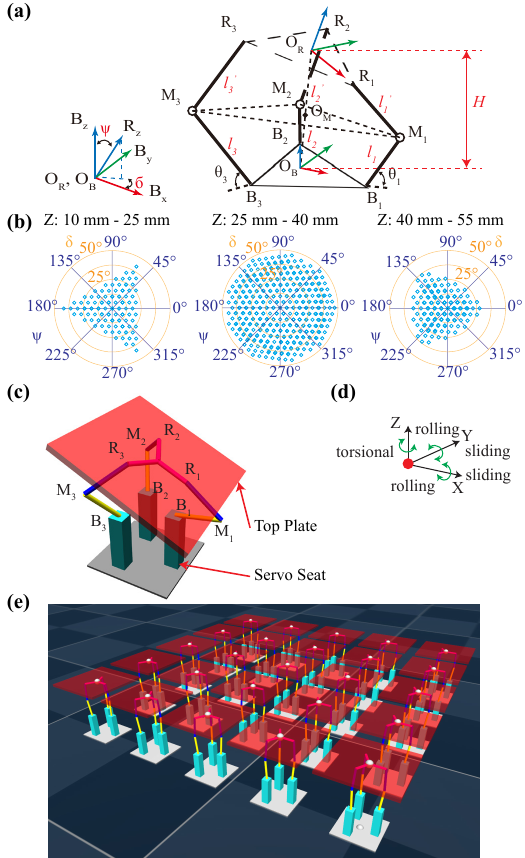}
    \caption{\textbf{Kinematic model, workspace, and simulation setups. }(a) Kinematic model of the Canfield origami structure; (b) non-monotonic behavior of end-effector's workspace from lower to higher Z-height configurations, first expanding then contracting; (c) single module model for simulation; (d) simulation contact model; (e) 5$\times$5 multi-module model for simulation.
    }
    \label{fig:modeling}
\end{figure}

This section discusses the kinematic modeling and workspace analysis of the Ori-Pixel platform, emphasizing its application in generating manipulation motion patterns. A dynamic model is then developed to simulate the dynamic behavior of the origami robot surface using MuJoCo~\cite{todorov2012mujoco}.

\subsection{Kinematic model and workspace analysis}
\label{sec:kinematic}
This work uses a 5$\times$5 grid of 25 3-DoF Canfield origami robots. Based on prior kinematic analyses~\cite{Canfield1998, 10122040}, the structure (\figureref{fig:modeling}(a)) consists of revolute joints $(B_1, B_2, B_3, R_1, R_2, R_3)$ and ball joints $(M_1, M_2, M_3)$. All linkages $(l_1, l_2, l_3, l_1^{'}, l_2^{'}, l_3^{'})$ are 30\,mm long, with $B_{1,2,3}$ and $R_{1,2,3}$ equidistant (20.21\,mm) from centers $O_B$ and $O_R$ respectively. Actuation angles ${\theta_i}, i\in{1,2,3}$ are determined from the top plate's pose parameters $(\delta, \psi, H)$ through:
\begin{equation}
\label{eq:kinematic}
    \theta_i=2\cdot arctan(t_i),  \theta_i \in[0, \frac{\pi}{2}],
\end{equation}
where:
\begin{equation}
    \begin{split}
        t_i &=\frac{-b_i\pm \sqrt{b_i^2-4a_ic_i}}{2a_i},\\
        a_i &=(r-l)(sin(\frac{\psi}{2})\cdot cos(\delta-\theta_i))-\frac{r_0}{2},\\
        b_i &=2l\cdot cos(\frac{\psi}{2}),\\
        c_i &=(r+l)(sin(\frac{\psi}{2})\cdot cos(\delta-\theta_i))-\frac{r_0}{2},\\
        r_0 &=\frac{H}{sin(\frac{\pi}{2}-\frac{\psi}{2})}.
 \end{split}
\end{equation}

The end-effector's inclination angle $(\psi)$ and height $(H)$ are key parameters constrained by the system's kinematics. As depicted in~\figureref{fig:modeling}(b), analysis of their workspace across three height configurations ($h\in[10, 25]$, $[25, 40]$ and $[40, 55]$ mm) reveals that the workspace first expands from lower to medium heights, then contracts at higher configurations. The workspace shows asymmetry across $\psi$ and $\delta$ ranges, necessitating separate optimizations for each direction of manipulations.

\subsection{Dynamic Simulation}
\label{sec:simulation}

The single-module model derived from the kinematic analysis is then developed for dynamic simulations in MuJoCo with a timestep of $5\times 10^{-4}$\,s using the default semi-implicit Euler integrator. As illustrated in \figureref{fig:modeling}(c), the revolute joints $B_1, B_2, B_3$ are connected to the base of each lower linkage, with their axes offset by $60$ degrees from one another. The ball joints $M_1, M_2, M_3$ link the lower linkages to the upper linkages, while the revolute joints $R_1, R_2, R_3$ connect the upper linkages to the top plate, sharing the same axis orientation as the joints $B_1, B_2, B_3$. A spring-damper model is applied to each joint, with a spring stiffness of $k_p = 0.2$ $N\cdot\text{m}/\text{rad}$ and a damping coefficient of $d = 0.1$ $N\cdot\text{m}\cdot\text{s}/\text{rad}$. The dimensions and masses of the linkages and top plates are derived from the same design parameters used in the prototype, as presented in~\cite{Oripixel}. Three motor actuators are implemented in position control mode with position feedback gain $k_p = 5$ and connected to the joints $B_1, B_2, B_3$. 

The single-module model is then replicated to form the 5$\times$5 module grid surface with identical distributions and dimensions as the prototype design presented in~\cite{Oripixel}. The multi-module MuJoCo model is depicted in~\figureref{fig:modeling}(e). 
This model includes 75 motor actuators, and all top plates feature contact models to simulate interactions with objects using soft contact dynamics with a solver tolerance of $10^{-6}$ and a maximum 30 iterations per timestep. As illustrated in~\figureref{fig:modeling}(d), the contact model incorporates sliding and rolling friction along the X and Y axes, and torsional friction along the Z axis. The sliding friction coefficient, $\mu_{slide}$, and the rolling friction coefficient, $\mu_{roll}$, are calibrated to $0.5$ and $0.01$, respectively. 

\section{CPG Motion Planning and Optimization for Manipulations}
\label{sec:motion_optimization}

\begin{figure*}[t]
    \centering
    \includegraphics[width = \textwidth]{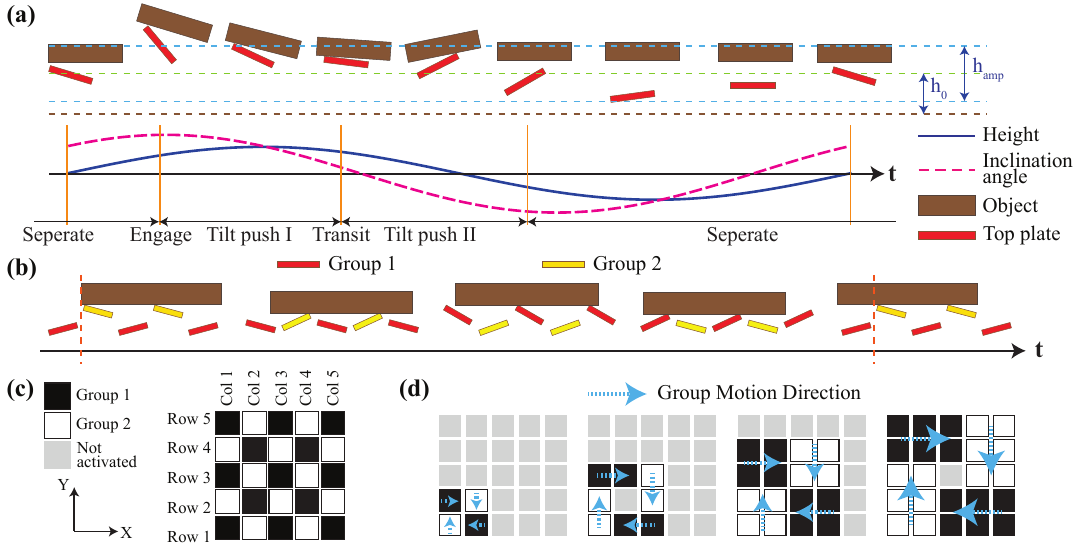}
    \caption{\textbf{Single-module CPG motion plan and inter-group motion plan. }(a) Single module motion plan; (b) multi-module manipulation motion plan. (c) inter-group motion planning for translation manipulations; (d) motion planning for clock-wise rotation manipulation.
    }
    \label{fig:CPG}
\end{figure*}

This section presents a novel CPG-based motion generation framework with simulation-based optimizations that, while demonstrated on Ori-Pixel platform, offer broad applicability across robotic systems. The proposed framework represents the first implementation of surface manipulations capable of handling diverse object geometries and stiffness. The following details the CPG parameter design, and the optimization framework across different manipulation modes.

\subsection{CPG-based Manipulation Motion Planning}
\label{sec:CPG_plan}
The CPG-based manipulation motion of a single 3-DoF Canfield origami robot mimics walking gait generation, consisting of three steps: object engagement through top plate lifting, object pushing through plate tilting, and plate retraction for disengagement, as shown in~\figureref{fig:CPG}(a). 

\begin{figure*}[t]
    \centering
    \includegraphics[width = \textwidth]{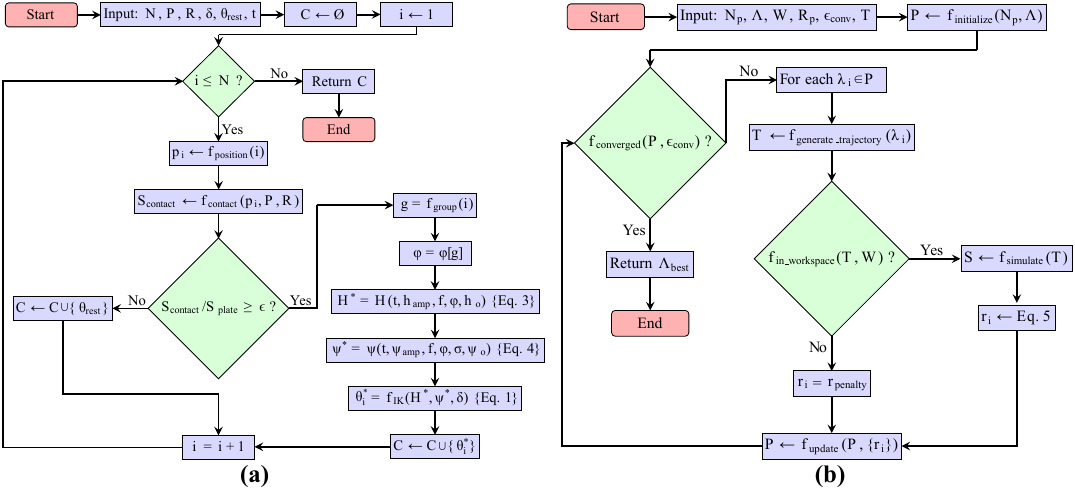}
    \caption{\textbf{Control and Optimization Frameworks.} (a) Control framework for CPG-based manipulation; (b) optimization framework for CPG parameters.
    }
    \label{fig:alg}
\end{figure*}

A CPG-based controller is designed to produce synchronized manipulation movements at the single module's top plate. These movements are governed by coupled sinusoidal functions in height $H$ and inclination angle $\psi$, as illustrated in~\figureref{fig:modeling}(a). The variation of $H$ and $\psi$ over time can be described as:
\begin{align}
    H(t) &= h_\text{amp} \sin(2\pi f \cdot t + \phi) + h_0\label{eq:height}, \\
    \psi(t) &= \psi_\text{amp} \sin(2\pi f \cdot t + \phi+ \sigma) + \psi_0\label{eq:psi},
\end{align}
where $h_\text{amp}$ denotes the amplitude of the height variation, $f$ is the frequency of motion, $\sigma$ represents the phase shift coupling between $H$ and $\psi$, $h_0$ and $\psi_0$ indicate the height and the inclination angle, respectively, of the top plate at its natural resting position, as depicted in~\figureref{fig:CPG}(a). Additionally, $\phi$ represents the inter-group phase shift used to coordinate multiple groups of modules for effective manipulations.

As shown in~\figureref{fig:CPG}(a), Eqs.~\eqref{eq:height} and \eqref{eq:psi} together define the motion pattern. The phase shift $\sigma$ determines the inclination angle $\psi$ when the top plate engages with the object, which dictates the manipulation direction. Specifically, if the first contact occurs during the tilt push I phase, where $\sigma \in [0, \pi)$, the manipulation direction is toward the left-hand side; if the first contact occurs during the tilt push II phase, where $\sigma \in [-\pi, 0)$, the direction shifts to the right-hand side. Furthermore, the phase shift $\sigma$ influences both the floating and tilting ranges of the object during manipulation, playing a key role in balancing the trade-off between speed and smoothness in the overall performance. This trade-off will be further studied during the optimization process in~\sectionref{sec:motion_plan}.

While single modules can complete manipulation cycles (\figureref{fig:CPG}(a)), additional module support during plate retraction is needed to prevent backward slippage. Thus, modules are divided into two groups using Eqs.~\eqref{eq:height} and \eqref{eq:psi} with identical parameters except for an inter-group phase shift $\phi$. As depicted in \figureref{fig:CPG}(b), this coordination enables simultaneous pushing and supporting motions for effective manipulations.

\subsection{Collective motion planning for manipulations using multi-module robotic surface}
\label{sec:motion_plan}
The inter-group motion plan developed for translational manipulations on the Ori-Pixel platform divides the modules into two diagonal groups, with their motion generated by the CPG described in~\sectionref{sec:CPG_plan}, as depicted in~\figureref{fig:CPG}(c). These groups are synchronized through the inter-group phase shift term, $\phi$. The diagonal symmetric configuration maintains object orientation during movement by applying forces without rotational torque, ensuring robust and stable motion.

The direction of translational manipulation is determined by two parameters: the azimuth angle $\delta$ (which defines the manipulation axis: $\delta = 0$\,degree for Y-axis and $\delta=90$\,degree for X-axis as shown in~\figureref{fig:modeling}(a) and \figureref{fig:CPG}(c)), and the in-group phase shift $\sigma$ (which determines direction along the chosen axis: $\sigma \in [0, \pi)$ for positive and $\sigma \in [-\pi, 0)$ for negative direction, as detailed in~\sectionref{sec:CPG_plan}). Together, these parameters enable omni-directional planar manipulation.

The rotational manipulation plan coordinates modules' translational movements in different directions. Objects must contact at least two by two top plates. Two module groups operate with $\phi=\pi$ phase shift. As shown in~\figureref{fig:CPG}(d), Group 1 moves along X-axis (positive in top-left, negative in bottom-right), while Group 2 moves along Y-axis (negative in top-right, positive in bottom-left) for clockwise rotation. Counter-clockwise rotation reverses these directions while maintaining $\phi$, ensuring continuous rotational manipulation.

In addition to the motion plans, the top plate's effective contact ratio ($S_{\textit{contact}} / S_{\textit{plate}}$) is crucial for manipulations, where $S_{\textit{contact}}$ represents the object-covered area and $S_{\textit{plate}}$ the total plate area. Modules are activated when this ratio exceeds a threshold $\epsilon$, otherwise returning to rest. This threshold, analyzed during optimization, ensures effective manipulation without obstruction. The complete control framework is illustrated in \figureref{fig:alg}(a).

\subsection{Optimization Framework for Manipulation Motions}
\label{sec:optimization}
A simulation-based optimization framework is developed to identify the optimal CPG parameters for different manipulation modes and objects, using the motion planned in~\sectionref{sec:motion_plan}. The complete process is elaborated in \figureref{fig:alg}(b).

The proposed framework utilizes the simulation model from Section~\ref{sec:simulation}, integrated with an evolutionary Bayesian hyperparameter optimizer~\cite{Cowen-Rivers2022-HEBO} to identify optimal parameter sets for different manipulation modes. The optimization search space spans eight parameters ${h_\text{amp}, \psi_\text{amp}, f, h_\text{0}, \psi_\text{0}, \phi, \delta, \epsilon}$ from Equations~\eqref{eq:height} and \eqref{eq:psi}. The amplitude and frequency ranges in Table~\ref{tab:opt_space} were bounded by our servo motors' physical limits. The initial height $h_0$ and orientation $\psi_0$ ranges were determined by the prototype's geometric design and kinematic workspace analysis as in~\sectionref{sec:kinematic}, while the phase parameters were constrained to ensure smooth transitions between motion states.
During the optimization process, the object is positioned at the center of one side of the robotic surface. The modules are commanded to move for 5\,seconds following the control protocol outlined in \figureref{fig:alg}(a), using the parameters suggested by the optimizer. The object's travel distance, Z-axis displacement, and rotation angles during the movement are evaluated using a cost function, which serves as reward feedback for the optimizer.

\begin{table}[htbp]
\begin{center}
\caption{Optimization Search Space}
\setlength{\extrarowheight}{1pt}
    \begin{tabular}{  c  c  c  c  p{5cm} }
    \hline
   \textbf{Parameter} & \textbf{Symbol} & \textbf{Search Space} & \textbf{Unit}\\
    \hline
    Height amplitude & $h_\text{amp}$ & $$[0.005, 0.04]$$ & m\\
    Inclination angle amplitude & $\psi_\text{amp}$ & $[0.35, 0.79]$ & radian\\
    Frequency & $f$ & $[0.1, 0.8]$ & Hz\\
    Resting height & $h_\text{0}$ & $[0.02, 0.04]$ & m\\
    Resting inclination angle & $\psi_\text{0}$ & $[-0.26, 0.26]$ & radian\\
    Height-inclination phase shift & $\sigma$ & $[0, \pi]$ or $[\pi, 2\pi]$ & radian\\
    Inter-group phase shift & $\phi$ & $[0, 2\pi]$ & radian\\
    Top plate contact threshold & $\epsilon$ & $[0.1, 0.5]$ & - \\
    \hline
    \end{tabular}
    \label{tab:opt_space}
\end{center}
\end{table}

A generalized cost function $J$ is constructed to assess manipulation performance, incorporating various manipulation objectives. As shown in Eq.~\eqref{eq:cost}, the cost function accounts for the object's absolute averaged translational speed $v$, the absolute averaged yaw speed $\omega$, the max roll ($\eta$) and pitch ($\rho$) angles, as well as the max displacement in z-direction throughout the manipulation process. The weights $\{\alpha, \beta, \gamma, \varsigma\}$ can be tuned to prioritize different manipulation objectives.

\begin{equation}
\begin{split}
J = &\ \alpha \cdot v + \beta \cdot \omega\\
    &\ + \gamma \cdot \left( \max_{t} \eta(t) + \max_{t} \rho(t) \right) + \varsigma \cdot \max_{t} z(t).
\label{eq:cost}
\end{split}
\end{equation}

For fast manipulations, where only the object's average manipulation speed $v$ is considered, the weights are set to $\{\alpha, \beta, \gamma, \varsigma \} = \{-1, 0, 0, 0\}$, ensuring that the optimizer focuses on maximizing the manipulation speed. In contrast, for smooth manipulations aimed at minimizing rotation, tilting, and shaking, the weights are adjusted to $\{\alpha, \beta, \gamma, \varsigma \} = \{-0.2, 0.3, 0.3, 0.3\}$, which directs the optimizer to prioritize reducing pose variations during the manipulation while still maintaining a reasonable speed. For rotational manipulations, the weights are set to $\{\alpha, \beta, \gamma, \varsigma \} = \{1, -1, 0, 0\}$, penalizing translational motions while rewarding yaw rotations. 

Using the optimization framework and simulation environments from~\sectionref{sec:simulation}, we optimized CPG parameters (\tableref{tab:opt_space}) for various manipulation motions (omnidirectional planar translations in fast/smooth modes and pure rotations). Due to the asymmetric workspace as discussed in \sectionref{sec:kinematic}, each case was optimized separately. For fast modes, we fixed $\phi=\pi$ to maximize contact time, while for smooth modes, $\phi$ was optimized to control contact transitions and minimize vertical displacement. The framework converged in 30 minutes on an AMD Ryzen Threadripper 7960X with 128GB RAM. Results are shown in supplementary video~1.

\section{CPG-Based Manipulations on Prototype}
\label{sec:prototype}

This section describes the laboratory experiments conducted with the Ori-Pixel robotic surface to manipulate various objects using the optimized CPG motions from~\sectionref{sec:motion_optimization}.

\begin{figure}[t]
    \centering
    \includegraphics[width = \columnwidth]{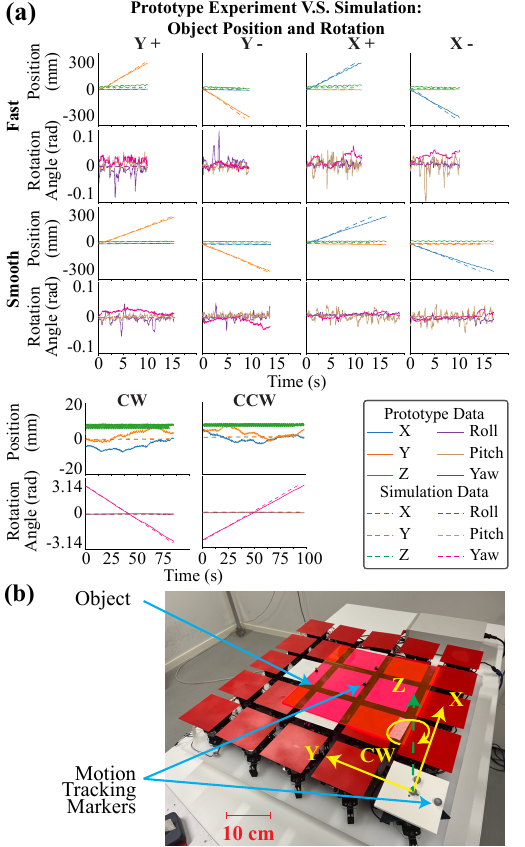}
    \caption{\textbf{Experimental validation.} (a) Comparison of simulated and prototype object motions for translational and rotational manipulations; (b) lab test setup.
    }
    \label{fig:prototype}
\end{figure}

\subsection{Experiment Setup}
\label{sec:exp_setup}
\begin{figure*}[t]
    \centering
    \includegraphics[width = \textwidth]{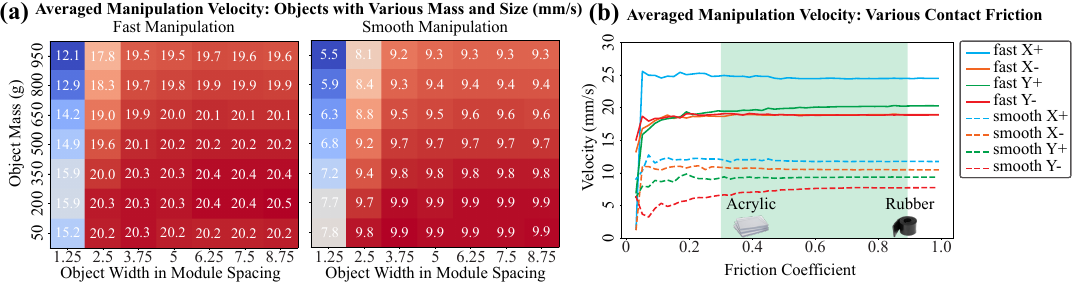}
    \caption{\textbf{Simulation results for system robustness analysis.} (a) Averaged manipulation velocity from simulation for various object masses and widths using the Ori-pixel module spacing of 120\,mm as a reference unit; (b) averaged manipulation velocity from simulation with various contact friction coefficients.
    }
    \label{fig:dis}
\end{figure*}
The Ori-Pixel robotic surface, comprising 25 modules each actuated by three Dynamixel XL-320 servos operating at 20\,Hz in position mode, is used for laboratory experiments (\figureref{fig:prototype}(b)). A Vicon Vero motion tracking system provides real-time object pose data for module activation control following \figureref{fig:alg}(a). Experiments are conducted with objects of varying size, weight, shape, and stiffness, as listed in \tableref{tab:exp}.

\subsection{CPG-Based Manipulation Experiments on Ori-pixel}
\label{sec:lab_exp}

The optimized CPG parameters are applied to manipulate a spectrum of objects as listed in~\tableref{tab:exp}. The system successfully manipulated objects ranging from a small plate (ii) to a much larger plate (iv), demonstrating the optimized motion's ability to handle a wide size range. The system also successfully manipulated objects with flexible materials and irregular shapes (v, vi) where position tracking becomes challenging. The platform operates in open-loop mode, with all modules activated following optimized CPG motions without position feedback. These experiments demonstrate that motions derived using the proposed framework can successfully handle objects of varying sizes, shapes, and stiffness, showcasing the robustness and versatility of the proposed manipulation strategy. Complete documentation is available in supplementary video~2.

\begin{table}[t]
\centering
\caption{Properties of Tested Objects}
\renewcommand{\arraystretch}{1.2}
\setlength{\tabcolsep}{4pt}
\begin{tabular}{c l l l c c}
\hline
\textbf{Index} & \textbf{Shape} & \textbf{Material} & \textbf{Size (mm)} & \textbf{Mass (g)} & \textbf{Tested Modes} \\
\hline
i   & Plate      & Acrylic  & 300$\times$300 & 254  & Fast, Smooth \\
ii  & Plate      & Wood  & 200$\times$200 & 172  & Fast, Smooth \\
iii & Cylinder   & Foam     & $\diameter$36$\times$140 & 9  & Smooth \\
iv  & Plate      & Acrylic  & 1000$\times$300 & 1000 & Fast \\
v   & Polo shirt & Fabric   & 400$\times$400 & 280  & Fast \\
vi  & Trilby hat & Straw   & 270$\times$250 & 55  & Fast \\
\hline
\end{tabular}
\label{tab:exp}
\end{table}

We evaluated the fast and smooth manipulation modes using object i as depicted in \figureref{fig:prototype}(a). Fast manipulation achieved higher velocities (30\,mm/s Y-direction, 25\,mm/s X-direction) compared to smooth manipulation (20\,mm/s Y-direction, 17\,mm/s X-direction). However, smooth manipulation demonstrated superior stability with lower Z-direction displacement (averaged standard deviation: 3.03\,mm vs 7.05\,mm) and rotation angles (averaged standard deviation: 0.0091\,rad vs 0.0133\,rad). To demonstrate stability, we successfully manipulated object i while supporting an unrestrained object iii (Figure~\ref{fig:concept}(b)(iii)). Pure rotational tests achieved average angular velocities of 0.079\,rad/s clockwise and 0.063\,rad/s counterclockwise. All experiments are documented in supplementary video~3.

\subsection{Sim-to-real Analysis}
\label{sec:sim_real}
To analyze the sim-to-real gap, we conducted simulations using object i with optimized CPG-based control parameters and compared them with experimental results, as shown in \figureref{fig:prototype}(a). Our simulation demonstrates strong alignment with the actual dynamic behavior during manipulation, though discrepancies were observed in rotation data during translation modes and position data during rotation modes. These gaps mainly stem from actuation delays between servos in the physical platform and natural variations in object placement during trials. Despite these differences, the control parameters optimized in simulation transferred effectively to real-world implementation, validating our sim-to-real approach.

\subsection{Conclusion}

This section evaluated the CPG-based manipulation motions derived from the simulation-based optimization process through prototype experiments. The experiments demonstrated high-fidelity sim-to-real transfer and validated the proposed framework by successfully manipulating objects of various size, shape, and stiffness, while executing fast and smooth manipulation modes to meet different performance requirements.
\section{Discussion}
\label{sec:discussion}
This section analyzes experimental results to evaluate our CPG-based manipulation framework. We demonstrate the framework's robustness across varying object properties (mass and size) and contact friction conditions, followed by a discussion of its key assumptions and limitations.

\subsection{Robustness Analysis}
\label{sec:robustness}
For robustness analysis of our proposed CPG-based manipulation framework, we conducted a series of simulations with the Ori-pixel platform serving as our experimental testbed. We analyzed box-shaped objects with a fixed height of 50\,mm but varying mass and width. The mass ranged from 50\,g to 950\,g in 150\,g increments, and using the Ori-pixel module spacing of 120\,mm as a reference unit, we varied object widths from 1.25 to 8.75 module spans in 1.25-span increments. We simulated translational manipulations in all directions using fast and smooth modes, with results shown in \figureref{fig:dis}(a). The analysis reveals that manipulation performance strongly correlates with module coverage, where objects spanning 2$\times$2 modules result in manipulation that is more sensitive to object properties and achieves lower velocities, while coverage of 3$\times$3 modules or more enables robust, high-velocity manipulation. This improved performance with larger coverage stems from better load distribution across modules, which helps mitigate the velocity reduction effects from increasing object mass. These results demonstrate the framework's robustness across a range of object masses and sizes.

We then investigated the effect of contact friction between the object and the platform. Using the same simulation setup with object i as in \tableref{tab:exp}, we varied the friction coefficient from 0.02 to 1 in 0.02 increments, testing all directional translational manipulations in both fast and smooth modes. The average velocities are shown in \figureref{fig:dis}(b). At friction coefficients below 0.3, manipulation velocity shows unstable saturation behavior. Above 0.3, the velocity stabilizes, indicating robust performance. The green shaded region (0.3-0.9) highlights that our proposed manipulation method works effectively with common materials ranging from acrylic (friction coefficient 0.4) to rubber (friction coefficient 0.9).

\subsection{Assumptions and Limitations}
\label{sec:sim_to_real}
The manipulation method presented here demonstrates robust performance across objects with diverse shapes, sizes, weights, and materials. For implementation on the current Ori-pixel platform, we assume objects have a flat contact surface, are larger than 150\,mm to effectively cover more than 2$\times$2 tiles and prevent falling into gaps between modules, and weigh less than 1500\,g due to actuator capabilities. 

As for limitations of the proposed framework, the robust manipulation performance requires friction coefficients above 0.3, though this encompasses most common materials from acrylic to rubber. Additionally, the framework has a resolution limitation requiring objects to span at least 2$\times$2 tiles to maintain consistent manipulation forces.
\section{Conclusion and future work}
\label{sec:conclusion}

This letter introduces a novel manipulation framework that uses a CPG-based motion generator to enable manipulation motions on a multi-module origami robotic surface. It also presents a simulation-based optimization method to find the best CPG parameters for various manipulation goals. The optimized manipulation motions are evaluated using both dynamic simulations and prototype experiments. This letter also showcases a series of demonstration experiments using the optimal CPG motions to manipulate objects of different sizes, shapes, and stiffness, highlighting robust and versatile manipulation across a wide range of objects.

In future work, reconfigurable module layouts will be investigated to enhance the platform's versatility. This improvement aims to expand the range of objects that can be effectively manipulated. Additionally, we plan to explore hybrid frameworks that combine learning-based methods with CPG to balance adaptability and control efficiency. Furthermore, we plan to explore hybrid frameworks that combine learning-based methods with CPG to enable more dynamic and complex manipulation tasks while maintaining control efficiency.

\newpage


\bibliographystyle{IEEEtran}


\end{document}